# AUTOMATED CONTENT GRADING USING MACHINE LEARNING

Theoretical Content Grading from Exam Papers

RAVINDER SAHARAN[1], RAHUL KR. CHAUHAN[2]

[1,2] dept. of Computer Science & Engineering
AMITY UNIVERSITY, Sector 125, Noida
Uttar Pradesh, India
[1] ravinder.saharan23@gmail.com; [2] rahulchauhan819@gmail.com

SIDDHARTHA SINGH[3], PRITI SHARMA[4]

[1] dept. of CSE; [2] Asst. Professor-II, dept. of CSE,
AMITY UNIVERSITY, Sector 125, Noida
Uttar Pradesh, India
[1] siddharthasingh321@gmail.com; [2] psharma11@amity.edu

*Abstract*— *Grading of examination papers is a hectic, time-labor intensive task and is often subjected to inefficiency and bias in checking. This research project is a primitive experiment in automation of grading of theoretical answers written in exams by students in technical courses which yet had continued to be human graded. In this paper, we show how algorithmic approach in machine learning can be used to automatically examine and grade theoretical content in exam answer papers. Bag of words, their vectors & centroids and a few semantic and lexical text features have been used overall. Machine learning models have been implemented on datasets manually built from exams given by graduating students enrolled in technical courses. These models have been compared to show the effectiveness of each model.*

## I. INTRODUCTION

All the major technical education systems aim at knowledge based evaluation, achieved through written examinations, where students give the test of their understanding and application of the domain. The grading and checking of these written examinations have certain challenges towards their automation and are therefore, limited to manual grading by human instructors. Since there is a vast number of domains and disciplines of which students are enrolled in, the challenge to even categorizing those disciplines into content-based and non-content based categories is even very strenuous.

Previous attempts to solve a few grading systems, such as Essay Grading (discussed in detail in section II) using machine learning, has been successful, but was only limited to linguistic domain. However, the automated grading of exam content in 'technical domain' can also be achieved to a decent extent using machine learning, which we discuss in this paper.

### A. Motivation

1.1 Many organizations and educational institutions have already introduced computer based tests, even for text based exams. Our project aims at solving the problem of automated grading of these tests. Also, text based content from hand written papers (after OCR to text processing), can similarly be checked and graded as well, although research on this is in progress.

It seems to many people that only human instructors can grade the papers as content in them is of various writing styles, sense, and linguistically difficult to be completely judged by a computer. While this might be true to some extent in general studies & courses, but in case of Technical domain and studies, e.g. Computer Science and Engineering, the most important features and determining features of the content are based on Bag of words, tf-idf patterns and some other text based features (classified). On top of these features, it's really important that the content is specific, unlike in 'Essay Grading' where the different thoughts of different writers is to be considered into account. Content in technical domain is specific, and requires strong machine learning algorithms with standardized dataset training, to obtain quality grading, because the grading in almost all colleges/institutions/universities is done on relative basis.

### B. Hypothesis & Challenges

The attempt to solve the problem of automated content grading in out perspective is based on the hypothesis and these points to be mentioned while attempting to present our research work:

- Technical domain: As many systems are developed already to solve general content, e.g. Grading of English essays, etc; our approach is to solve the content in technical domain, which had not yet been successfully attempted to be solved. Technical domain would include courses like Bachelor of Science, Master of Science, Bach. Of Technology &

- Engineering, Arts, etc. and not General English based, e.g. Psychology, Law etc.
- Computer Based Tests: The exams of whom the content is to be graded by this approach will give us text input through the computer based examination. Many organizations today are adopting computer based exams.
- Data collected: Only the "theoretical content" can be graded using this approach is to be graded, as it takes about 75-85% of total time for a human grader to check and grade a paper for only the theoretical content. Checking of other types of content, e.g. Illustrative, diagrammatic, numeric etc. takes a very jiffy amount of time as compared to theoretical content. As a result, we can achieve significant amount of time being reduced while checking the theoretical content i.e. an expected 60-80% reduction in time spent by a human grader as compared to a computer grading system.
- Large collection of data: As the content to be checked for human graders is generally large, it is expected that this system be implemented on domains with huge data content to be processed.

*C. Software tools used:*

We chose to implement our model in Python 2.7.x, as there is a vast set of libraries for working with natural language processing. We have used the Natural Language Toolkit (NLTK) and *textmining* for most NLP tasks. Other libraries: numpy, scipy, xlrd, xlwt, word2vec etc. have been used for various tasks.

## II. LITERATURE OVERVIEW

There are many works related to machine learning with the intent to automate grading systems of essay writing systems. Some of the works related to it are:

*A. "Automated Essay Grading using Machine Learning"*
    - Mahana, Jons et. al. -Dec 2012 [1]
"The project aims to build an automated essay scoring system using a data set of 13000 essays from kaggle.com. These essays were divided into 8 different sets based on context. We extracted features such as total word count per essay, sentence count, number of long words, part of speech counts etc from the training set essays. We used a linear regression model to learn from these features and generate parameters for testing and validation. We used 5-fold cross validation to train and test our model rigorously. Further, we used a forward feature selection algorithm to arrive at a combination of features that gives the best score prediction. Quadratic Weighted Kappa, which measures agreement between predicted scores and human scores, was used as an error metric. Our final model was able to achieve a kappa score of 0.73 across all 8 essay sets. We also got a good insight into what kind of features could improve our model, for example N-Grams and content testing features."

*B. "Automated Essay scoring using machine learning"*
    Song, Zhao –Mach. Learning Sessions, Stanford University, 2012 [2]
"We built an automated essay scoring system to score approximately 13,000 essay from an online Machine Learning competition Kaggle.com. There are 8 different essay topics and as such, the essays were divided into 8 sets which differed significantly in their responses to the our features and evaluation. Our focus for this essay grading was the style of the essay, which is an extension on the studies conducted determining the quality of scientific articles by adding maturity to the feature set (Louis and Nenkova, 2013). An aspect of this project was to recognize the difference between the advanced nature of scientific articles to the coherency of middle to high school test essays. We evaluated Linear Regression, Regression Tree, Linear Discriminant Analysis, and Support Vector Machines on our features and discovered that Regression Trees achieved the best results with k = 0:52."

*C. "Automated Essay Scoring"*
    Murray, et. al. - 2012 [3]
"Analyzing natural language, or free-form text used in everyday human-to-human communications, is a vast and complex problem for computers regardless of the medium chosen, be it verbal communications, writing, or reading. Ambiguities in language and the lack of one "correct" solution to any given communication task make grading, evaluating or scoring a challenging undertaking. In general, this is a perfect domain for the application of machine learning techniques with large feature spaces, and huge amounts of data containing interesting patterns. In this project, we explore the use of linear regression from text features to directly predict the score of a given essay. Using l1 regularization, we take a large feature space consisting of a variety of linguistic features and determine the most predictive ones. We are able to significantly reduce prediction error and obtain state-of-the-art results, comparable to human annotators."

*D. "Automated Essay Scoring (Swedish)"*
    Östling, Robert, et al. - USA (2013) [4]
"The AES systems were based on standard supervised machine learning software, i.e., LDAC, SVM with RBF kernel, polynomial kernel and Extremely Randomized Trees. The training data consisted of 1500 high school essays that had been scored by the students' teachers and blind raters. To evaluate the AES systems, the agreement between blind raters' scores and AES scores was compared to agreement between blind raters' and teacher scores. On average, the agreement between blind raters and the AES systems was better than between blind raters and teachers. The AES based on LDAC software had the best agreement with a quadratic weighted kappa value of 0.475. In comparison, the teachers and blind raters had a value of 0.391. However the AES results do not meet the required minimum agreement of a quadratic weighted kappa of 0.7 as defined by the US based non-profit organization Educational Testing Services."

E. *"Enriching Automated essay scoring using Discourse Marking"*

    *Burstein, et. al. - ETS, NY and Hunter college, NY (2001)* [5]

"Electronic Essay Rater (e-rater) is a prototype automated essay scoring system built at Educational Testing Service (ETS) that uses discourse marking, in addition to syntactic information and topical content vector analyses to automatically assign essay scores. This paper gives a general description of e-rater as a whole, but its emphasis is on the importance of discourse marking and argument partitioning for annotating the argument structure of an essay. We show comparisons between two content vector analysis programs used to predict scores, EssayContent and ArgContent. EssayContent assigns scores to essays by using a standard cosine correlation that treats the essay like a "bag of words," in that it does not consider word order. ArgContent employs a novel content vector analysis approach for score assignment based on the individual arguments in an essay. The average agreement between ArgContent scores and human rater scores is 82%, as compared to 69% agreement between EssayContent and the human raters. These results suggest that discourse marking enriches e-rater's scoring capability. When e-rater uses its whole set of predictive features, agreement with human rater scores ranges from 87% - 94% across the 15 sets of essay responses used in this study."

## III. DATASET

The Dataset required for this experiment needed to be from one of the technical courses in which student are enrolled. The standard exam answer papers were taken from the mid-term (minor) examinations, since we could not directly get the major examination answer sheets because of some administrative issues. We collected theoretical answers, and to begin with, 3 subjects were chosen. These were all technical writing subjects. A total of 4 datasets were made, manually by our team. Dataset contained their answer in one column, along with its human-evaluated score in the other column. We split the datasets into 70 – 30 scheme of training and testing. We managed to work on 350 data entries in all sets sets included. This seems quite less as compared to what machine learning demands, but Firstly, since our feature model works on neural network unsupervised vector encoding & Secondly, because our problem domain is 'technical', unlike in Essay Grading, this dataset is enough to put up the capabilities of our approach, to a considerable extent.

## IV. METHODOLOGY

Our methodology comprises of standard training and testing machine learning approach. For this, feature vector -

Fig. Methodology & Overview of Flow (split in 2 halves)

- generation and training, score prediction after testing, and comparisons based on the models used, is described in the following Flow Diagram, followed by section wise discussion of these steps:

### A. Features

As discussed in section 1.2, the features required for solution of our problem would be very different from those used for Essay Grading approaches, or any other NLP-intensive feature generation approach.

As discussed, we modeled using python, we used NLTK and tools from this library available in python for pre-processing of text. After that we processed it into vectors, using word2vec implementation in python.

- Bag of Words(BoW): BoW lists words with their word counts per document. In the table where the words and documents effectively become vectors are stored, each row is a word, each column is a document and each cell is a word count. Columns (of equal length) represents each document in the corpus. Those are wordcount vectors, an output stripped of context.

  Vectors of words before being fed into the neural-net are normalized and combined. Hence, frequencies of occurrence of words are converted into probabilities w.r.t. the document. Levels of probabilities in the neural net will determine their weights and ranks.

- Tf-IDF (Term frequency-inverse document frequency): It measures the term frequency of a given word in a document. But because words such as "and" or "the" appear frequently in all documents, those are systematically discounted. That's the inverse-document frequency part. The more documents a word appears in, the less valuable that word is as a signal. That's intended to leave only the frequent AND distinctive words as markers. Each word's TF-IDF relevance is a normalized data format that also adds up to one.

$$w_{i,j} = tf_{i,j} \times \log\left(\frac{N}{df_i}\right)$$

$tf_{i,j}$ = number of occurrences of $i$ in $j$
$df_i$ = number of documents containing $i$
$N$ = total number of documents

- "Those marker words are then fed to the neural net as features to determine the topic covered by the document that contains them. While simple, TF-IDF is incredibly powerful, and contributes to such ubiquitous and useful tools as Google search." [6]

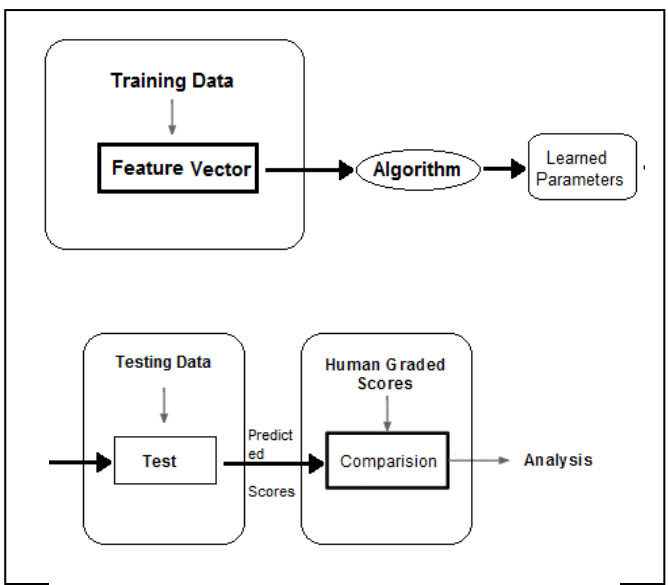

- Skip-grams: In the skip-gram model, used in Word2vec [7], a neural network is trained with intent to perform tasks, a single hidden layer is used, without actually using that neural network. The objective here is to learn the hidden layer content weights, which are actually "word vectors" that are being learnt.

These features form the basis for the vectors, used for training of the dataset and they act as proxies for the grading of content as if it were a human grader's basis for scoring the content.

Till now, it has only been preprocessed text, from the dataset, ready to be converted to vectors on the basis of these above features of our feature model.

### B. Training Algorithms

Algorithms that are used for training and score prediction include Random forest algorithm, Hierarchical softmax algorithm (used in Word2vec package library).

- Random Forest Algorithm: It is a versatile algorithm capable of performing different tasks such as regression, classification tasks, etc. for training and score prediction. It facilitates down-up formation of a powerful model combined from various weak models, therefore it is an association type learning model. It takes into account methods that facilitate dimension reduction, outlying values as well as missing links & other steps of data exploration, and does quite a great job.

    It works on formation of various trees as opposed to other models, such as CART Model [8]. The task of classification of a new object is based on each tree's classification on the ground of attributes set, and is said to 'vote' for that class. Votes term here signifies the majority that determines the classification. The task of regression takes the trees' average output.

    Several advantages of why random forest is great for such tasks as the one we intent to do, include the powerful high-dimension handling of large amounts of data with ability to handle several thousands of variables, estimation capabilities of missing data, and methods for balancing errors in datasets, especially unlabeled data. [9]

- Hierarchical softmax Algorithm (default training algorithm in Word2vec): This is inspired by and an approximate method to binary tree method that was proposed by Morin and Bengio [10]. Leaves replace the conventional softmax layer with a hierarchical layer, facilitating sequence of probabilistic calculations instead of separate calculations for each word, saving the normalization calculation over all words.

    "Hierarchical softmax provides for an improvement in training efficiency since the output vector is determined by a tree-like traversal of the network layers; a given training sample only has to evaluate /update **O(log(N))** network units, not **O(N).** This essentially expands the weights to support a large vocabulary - a given word is related to fewer neurons and visa-versa." [11]

    Hierarchical softmax is the default method for sampling training data using the skip-gram model that we used (as mentioned in part A of this section).

## V. EXPERIMENTAL RESULTS & DISCUSSION

### A. Individual observations

Predictive scores were found for distinct data sets and were compared to human graded scores, using 3 models namely, "Bag of words", "Bag of vectors" and "Bag of centroids" [12].

- Part-1 observations:
    - This is Result of From the TestData when observed w.r.t. Trained Data.
    - Only Keywords are extracted and then data is Trained via Random Forest Algorithm.
    - Labels: Red are predicted scores and Blues are given scores (testing set). Across x are different questions, and across y are marks

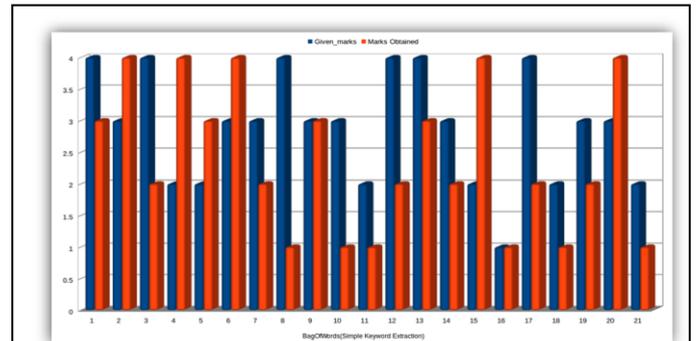

Fig: Marks distribution using the model of "Bag of words"

- Part-2 observations:
    - This is Result of from the when observed w.r.t. Trained Data.
    - In this an unlabeled dataset is trained via a trained dataset then this new data is then compared with new TestDataSet.

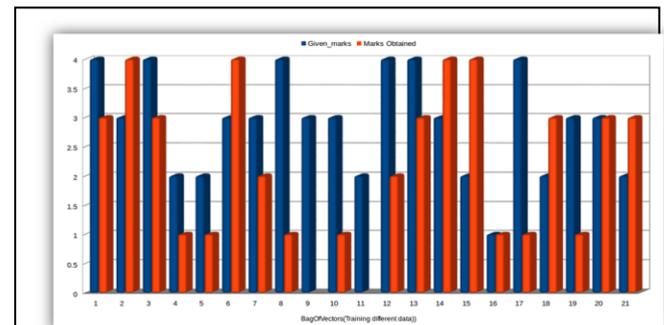

Fig: Marks distribution using the model of "Bag of vectors".

- Part-3 observations:
  - This is Result of from the when observed w.r.t. Trained Data, in which the data is clustered, with reference to the meanings of words. Through this, 9 clusters were formed and the results of the same were compared with TrainingDataSet.

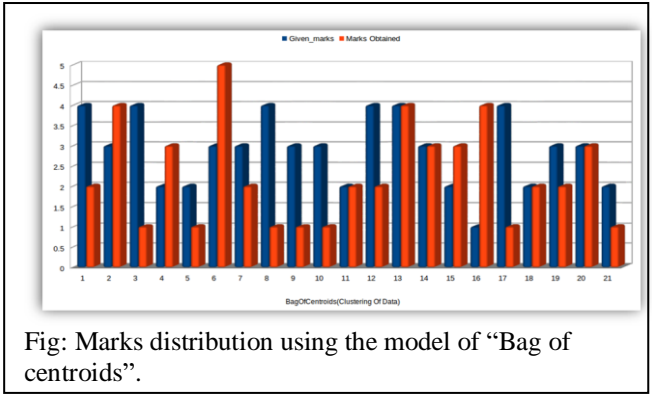
Fig: Marks distribution using the model of "Bag of centroids".

Note that the equation is centered using a center tab stop. Be sure that the symbols in your equation have been defined before or immediately following the equation. Use "(1)," not "Eq. (1)" or "equation (1)," except at the beginning of a sentence: "Equation (1) is ..."

B. *Comparisions*

The three models were compared w.r.t. each dataset, and the results are graphically represented as multiple bar charts as below:
*Note: These comparisons are across all the three methods as tested in part A of this section.*

- Dataset 1 comparisions:

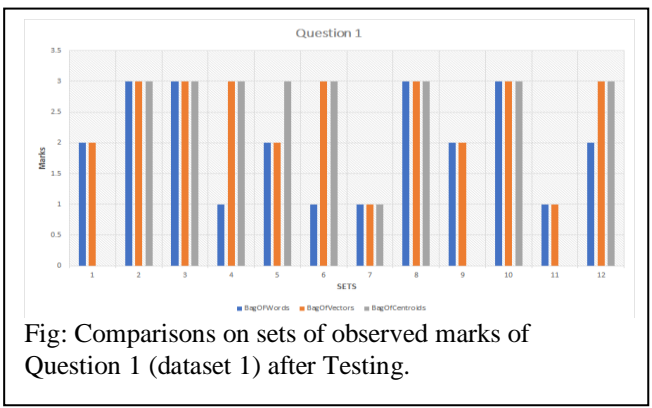
Fig: Comparisons on sets of observed marks of Question 1 (dataset 1) after Testing.

- Dataset 2 comparisions:

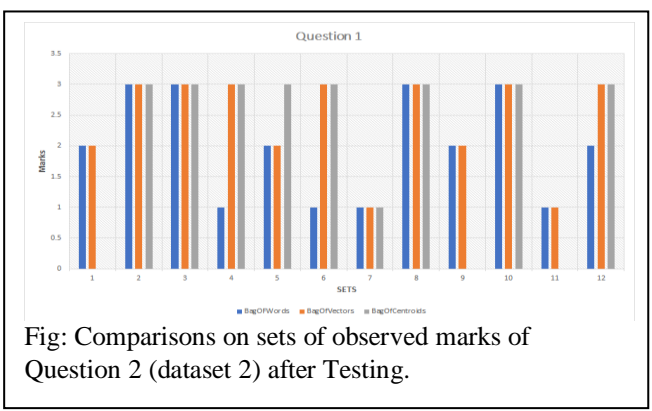
Fig: Comparisons on sets of observed marks of Question 2 (dataset 2) after Testing.

- Dataset 3 comparisions:

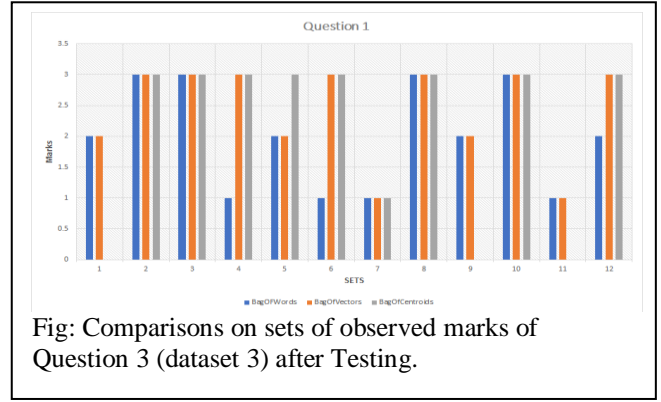
Fig: Comparisons on sets of observed marks of Question 3 (dataset 3) after Testing.

As we can observe from the charts, the most close and accurate results are observed in the 3rd model across all the datasets. The reason is that it takes "centroids of words" as the vector representation of the features. This works well with the Random forest algorithm to form individual trees for the clusters of data, which is done using K-means clustering[13]. This gives more accurate probabilistic values for the words (as described in the section IV-B, Random forest algorithm description).

C. *Error Analysis*

We calculated the weighed kappa for each set, and is plotted as follows w.r.t. each set:

| Sets | Weighed Kappa |
|---|---|
| Set 1 | 0.464 |
| Set 2 | 0.553 |
| Set 3 | 0.613 |
| **Average** | **5.5** |

Table: Error metric comparisons.

Although we have seen that the 3rd model gave us the best results, we have tabulated the kappa metrics as error analysis, and its clearly observed that despite the challenges (as discussed in the next section) we have observed a decent accuracy in the agreement of the human graded scores and the machine learned scores.

*D. Challenges & limitations for optimal results*

- Dataset limitations: The dataset for the major exams which could provide us all the numbers and stats of the amount of training and testing data that we need, was unfortunately not available. The dataset manually collected by us was from the minor exam papers of students from B.Tech. courses.

- Types of content in different questions: The aim of every question in exams is not same. For example: Essay type questions, descriptive questions, logic based questions, short logical, reasoning based etc. The list is too long. This is included in our future research for the feature models.

- Priority of content: The priority would have been higher which would certainly have improved our dataset quality. Students are always the most serious while giving major exams (end term examinations) rather than the mid-term examinations.

## VI. DISCUSSIONS & CONCLUSION

*A. Discussion*

Analysis of Observations according to the models implemented on all the datasets (or Questions; each dataset is based on the answers of different questions):

| Ques | Characteristic | Observations & Analysis |
|---|---|---|
| Q1 | The question is a moderately elaborative and a mild variety of styles and vocabulary is expected. | Results of the models are comparable to the human graded results. The clustering model gives the closest grades. |
| Q2 | The question is very comprehensive type, and the answers are descriptive. This allows the answers to contain more diverse vocabulary, more comprehensive answers, and a wide range of writing styles. A wide variation is also expected in the answers written by the students. | Most closely accurate results, the clustering model gives the most apt observations, where the graded marks of the system are comparable to the human graded marks on a good degree of comparison. This shows that the grading of most theoretical content type questions can be closely accurate on a Relative grading scale. |
| Q3 | The question is direct logic based, which indicates that the answers would contain a less overall vocabulary, limited writing style and low variations in answers/content. | The results are quite surprising; according to the attempt percentage, the 3rd model shows that the results are very strict according to human graded scored, and this can be limited using some other custom feature models to improve efficiency in such cases. |

*Table: Observations Analysis of all three Questions.*

*B. Conclusion*

We have identified that the type of data optimal for theoretical content grading using machine learning. We observed by the experiment results that the second dataset which contained descriptive theoretical question, gave us the best results so far. So we can conclude that the most descriptive, essay type questions (atleast 150 words, upto a prescribed limit) and those with detailed explanations and definitions are best type of content.

*C. Future Scope and Proposed Enhancements*

➢ While the deep exploration and addressing of the acquisition as a result of students' deep ideology has possibility of extensions and improvements, it can also be archaic.

➢ This paper has also represented how content grading in a big-data based technical domain can also be solved using this approach. However, the field statistical mathematical machine learning also has increasing and improving data models. The field is deep and there are promising new ways to think about it. But this paper has explained how new and simple model of visual words and significant is a vast area of exploration.

➢ However, more grammatical and vocabulary challenges are being faced in this type of tokenized machine learning, and is a great area of research and also has broader application to newer educational challenges.

➢ In future work, we plan to establish in-depth learning including grammar and spelling checking algorithms so that the result derived is more accurate and solid.

➢ The negative sampling approach is however not as fast in execution, but is definitely better in results and therefore is open to implementation over and above some exceptional feature models for unique datasets.